\documentclass{article}

\usepackage{PRIMEarxiv}

\usepackage[utf8]{inputenc} 
\usepackage[T1]{fontenc}    
\usepackage{hyperref}       
\usepackage{url}            
\usepackage{booktabs}       
\usepackage{amsfonts}       
\usepackage{nicefrac}       
\usepackage{microtype}      
\usepackage{lipsum}
\usepackage{fancyhdr}       
\usepackage{graphicx}       
\graphicspath{{media/}}     

\title{Optimal Decision Making Through Scenario Simulations Using Large Language Models
}

\author{
  Sumedh Rasal \\
  Georgia Institute of Technology \\
  Chicago, IL\\
  \texttt{srasal3@gatech.edu} \\
  \And
  E. J. Hauer \\
  \texttt{EJ@ThoughtCrafters.com} \\
}

\begin{document}
\maketitle

\begin{abstract}
The rapid evolution of Large Language Models (LLMs) has markedly expanded their application across diverse domains, transforming how complex problems are approached and solved. Initially conceived to predict subsequent words in texts, these models have transcended their original design to comprehend and respond to the underlying contexts of queries. Today, LLMs routinely perform tasks that once seemed formidable, such as writing essays, poems, stories, and even developing software code. As their capabilities continue to grow, so too do the expectations of their performance in even more sophisticated domains.

Despite these advancements, LLMs still encounter significant challenges, particularly in scenarios requiring intricate decision-making, such as planning trips or choosing among multiple viable options. These tasks often demand a nuanced understanding of various outcomes and the ability to predict the consequences of different choices, which are currently outside the typical operational scope of LLMs.

This paper proposes an innovative approach to bridge this capability gap. By enabling LLMs to request multiple potential options and their respective parameters from users, our system introduces a dynamic framework that integrates an optimization function within the decision-making process. This function is designed to analyze the provided options, simulate potential outcomes, and determine the most advantageous solution based on a set of predefined criteria. By harnessing this methodology, LLMs can offer tailored, optimal solutions to complex, multi-variable problems, significantly enhancing their utility and effectiveness in real-world applications. This approach not only expands the functional envelope of LLMs but also paves the way for more autonomous and intelligent systems capable of supporting sophisticated decision-making tasks.

\end{abstract}


\section{Introduction}
Large language models (LLMs) have demonstrated remarkable utility across various domains \cite{brown2020language} \cite{thoppilan2022lamda}, showcasing their ability to solve programming challenges \cite{chen2021evaluating}, elucidate complex topics in simpler and more understandable language, and generate unique texts that include novel and interesting content \cite{ouyang2022training} \cite{wei2022emergent} \cite{bubeck2023sparks}. From generating human-level text to tackling intricate programming questions, LLMs have continually proven their versatility and effectiveness \cite{saunders2022self} \cite{touvron2023llama} \cite{devlin2018bert} \cite{openai2023gpt}.

Despite their impressive capabilities, LLMs face significant limitations when addressing complex problems that require more than just pattern recognition and language generation \cite{liu2023your}. While LLMs can autonomously operate and learn patterns to generate coherent text, they are not inherently designed to tackle complex, multifaceted problems that involve multiple possible solutions \cite{rajani2019explain} \cite{ling2017program} \cite{cobbe2021training} \cite{chen2019neural} \cite{patil2023gorilla} \cite{lu2023chameleon} \cite{yao2022react} \cite{li2023camel}. For instance, solving programming problems may seem magical at first glance, but it relies on the LLM's ability to recognize and apply patterns, structures, and meanings inherent in the problem statements. Given sufficient data for training, the architecture of LLMs is adept at understanding these patterns and producing accurate solutions.

However, when tasked with solving design problems—where multiple potential solutions exist—the responses from LLMs tend to be generic and lack the specificity needed to identify the optimal solution \cite{sap2022neural}. These answers are challenging to quantify, as many possible right answers exist. This raises the question: how can we enable LLMs to effectively tackle such problems? \cite{liang2023encouraging} \cite{qian2023communicative} \cite{lazaridou2020multi} \cite{wu2023large} \cite{rasal2024llm}

We propose an innovative approach that empowers LLMs to evaluate multiple potential options for a given problem and either provide the user with the best possible solution or offer insights into the probability or likelihood of each solution's success. Our method involves leveraging the LLM's ability to simulate various scenarios, thereby enhancing its decision-making capabilities \cite{schick2023toolformer}.

To achieve this goal, it is essential to equip LLMs with the necessary tools to strategize and think critically about complex problems. Rather than constantly retraining LLMs with new data or problem statements, our approach allows LLMs to run multiple simulations of a given problem statement \cite{susskind2000using}. This enables the model to provide users with the best approach and the likelihood of success based on prior information and the outcomes of these simulations \cite{rasal2024artificial}.

By integrating this method, we aim to significantly expand the scope of LLM applications, allowing them to address more complex and dynamic problems effectively. This paper outlines our proposed approach, discusses its implementation, and explores its potential impact on enhancing the capabilities of LLMs in solving multifaceted problems.

This is what we propose:
\begin{itemize}
\item A framework that enables LLMs to run multiple simulations of a given problem statement.
\item Ability to incorporate the framework seamlessly within the existing LLM architecture without the need for constant retraining.
\item Utilize prior information and simulation results to provide probabilistic insights and optimal solutions to complex, multifaceted problems.
\end{itemize}

\section{Methodology}
Recently, large language models (LLMs) have been employed in numerous applications. For example, they are used to summarize text, generate ideas, and solve novel coding problems. They can be prompted to assume various personas to answer questions in specific contexts, such as legal, business, or sports. LLMs are capable of holding extended conversations, though they have their shortcomings. One key limitation is their dependency on pre-existing knowledge bases for retrieving factual information, which can lead to inaccuracies when faced with novel questions, often resulting in random or incoherent text, a phenomenon known as hallucination \cite{azamfirei2023large}.

Retraining LLMs \cite{rasal2024beyond} to expand their knowledge base is an expensive and resource-intensive process. While creating context-specific LLMs is a viable alternative, it does not solve the problem of developing an all-purpose LLM. Moreover, the concept of an all-purpose LLM leading to true artificial general intelligence (AGI) remains questionable. Without perceptors, an LLM cannot achieve true AGI \cite{rasal2024navigating}. Therefore, we propose an approach that uses LLMs as the core of the operation, augmented by additional perceptors to aid in decision-making and provide users with solutions to complex problems.

Our system architecture comprises the following components: User Input Interface, LLM Chat Agent, Simulation Module, Optimization Engine, Context-Aware Data Warehouse, and Result Interface. We will now discuss each component to understand the overall workflow.

\textbf{User Input Interface}

This module serves as the portal through which users submit their problem statements. It is powered by an LLM chat agent, which extracts user requirements and converts them into input parameters for the simulation module. The chat agent also queries context-aware data relevant to the user problem and asks pertinent questions to ensure all necessary information is gathered. This enables users to make informed decisions once the results are computed. Our approach is designed to tackle problems that resemble decision tree scenarios or combinatorial problems.

\textbf{LLM Chat Agent}

We utilize OpenAI's GPT-4 chat agent with specific prompts to engage with users and extract detailed information about their problem statements. These details are crucial for the simulation module to run multiple scenarios on the given problem. The optimization engine then narrows down the best options for the user, considering the constraints provided. The chat agent communicates with the optimization module and the data warehouse to ensure all variables are captured and conveyed accurately.

\textbf{Simulation Module}

This module employs a Python-based Monte Carlo simulation package. The chat agent translates the user’s problem statement into an optimization problem and runs numerous simulations. The simulation module is responsible for selecting the appropriate optimization approach for the user’s problem. The results from this module are then used by the optimization engine to generate a function that will be executed. Our architecture is designed to be flexible, allowing the use of different programming languages for the simulation and optimization processes.

\textbf{Optimization Engine}

The optimization engine, also based in Python, analyzes the results from the simulation module and provides a solution to the user’s problem statement. This engine ensures that the data queried from the data warehouse is incorporated into the output, offering users a comprehensive view of the solution, including how each variable impacts the overall result.

\textbf{Feedback Module}

Our architecture includes a feedback module that captures the user’s problem statement, the output provided, the optimization formula, and the simulation details. Users can also provide feedback on the effectiveness of the system’s output. All this data is stored in a relational database for potential retraining and future improvements.

This methodology leverages the strengths of LLMs while addressing their limitations by integrating additional components to handle complex, multi-variable problems. By running multiple simulations and utilizing an optimization engine, we provide users with robust, data-driven solutions that are tailored to their specific needs.








\section{Experiments}
To illustrate the practical application of our proposed system, we conducted an experiment focusing on the decision-making process involved in choosing between buying and leasing a car. This experiment demonstrates how the various modules in our system work together to provide users with optimal solutions based on their specific needs and constraints. We will walk through the experiment step-by-step, highlighting the interactions between the user and the LLM agent, as well as between the LLM agent and the Python interpreter.

\textbf{User Input Interface}

The user begins by interacting with the User Input Interface, which is powered by the LLM chat agent. The user provides their initial query about whether to buy or lease a car.

\begin{verbatim}
User: "I'm trying to decide if I should buy or lease a car. Can you help me figure out 
the best option?"

LLM Agent: "Of course! To help you decide, I'll need some additional information. Can 
you tell me about your driving habits, budget, and any preferences you have for the car?"

User: "I drive about 15,000 miles a year. My budget for monthly payments is around $400. 
I prefer a new car with good fuel efficiency."
\end{verbatim}

The LLM agent gathers these inputs and converts them into specific parameters for the simulation module.

\textbf{LLM Chat Agent}

The LLM chat agent continues to engage with the user to extract more detailed information and ensure all necessary variables are captured.

\begin{verbatim}
LLM Agent: "Great! I need to know the following details to proceed: the expected duration 
of ownership if you choose to buy, any down payment you can make, and your estimated 
annual maintenance costs. For leasing, I need to know the lease term you're considering 
and any potential mileage overage charges."

User: "If I buy, I plan to keep the car for about 5 years. I can make a down payment of 
$3,000, and I estimate annual maintenance costs at $500. For leasing, I’m considering a 
3-year term with an allowance of 12,000 miles per year, and the overage charge is 15 
cents per mile."
\end{verbatim}

The LLM agent now has all the required information and prepares to communicate with the simulation module.

\textbf{Simulation Module}

The LLM agent converts the user’s problem statement into an optimization problem and sends these parameters to the Python-based Monte Carlo simulation module.

\begin{verbatim}
LLM Agent (to Simulation Module): "Run simulations to compare the total cost of ownership 
(TCO) for buying versus leasing a car with the following parameters: 15,000 miles/year, 
$400/month budget, $3,000 down payment for buying, 5-year ownership, $500 annual 
maintenance for buying, 3-year lease term, 12,000 miles/year allowance, and 15 
cents/mile overage charge."

Simulation Module: "We will use a Monte Carlo simulation strategy. This strategy was 
selected because it allows us to model the probability of different outcomes in a process 
that cannot easily be predicted due to the intervention of random variables. It is 
particularly useful for complex scenarios where multiple variables can impact the 
outcome."
\end{verbatim}

The simulation module runs multiple scenarios to estimate the TCO for both buying and leasing the car over the specified periods.

\textbf{Optimization Engine}

The results from the simulation module are analyzed by the optimization engine. The engine calculates the TCO for both options and identifies the most cost-effective solution based on the user’s inputs and constraints.

\begin{verbatim}
Simulation Module (to Optimization Engine): "Here are the results: TCO for buying over 
5 years is $25,000, including down payment, monthly payments, and maintenance. TCO for 
leasing over 3 years, considering mileage overage, is $14,000. Projected costs for the 
next lease term are $14,000, making a 6-year lease TCO approximately $28,000."

Optimization Engine: "Based on the simulations, buying the car is more cost-effective 
than leasing for 6 years."
\end{verbatim}

\textbf{Result Interface}

The LLM agent communicates the final results to the user through the Result Interface.

\begin{verbatim}
LLM Agent (to User): "After analyzing your inputs, the total cost of ownership (TCO) for 
buying the car over 5 years is $25,000. The TCO for leasing over 6 years is approximately 
$28,000. Therefore, buying the car is the more cost-effective option for you."

LLM Agent: "To provide more detailed insights, here is the breakdown of the 
probabilities affecting each variable:

- Down payment: Fixed at $3,000
- Annual maintenance: Estimated $500 with a variability of ±$100, contributing 5% to the 
overall TCO uncertainty
- Mileage overage charge: $0.15 per mile for miles over 12,000/year, contributing 10% to 
the leasing TCO variability
- Monthly payment: Fixed at $400, with a potential variation of ±$50, contributing 3% to 
the overall TCO uncertainty

Thus, the expression of the total cost of ownership (TCO) is:
\end{verbatim}

\( TCO = Down Payment + \sum (Monthly Payment \times Months) + \sum (Maintenance Cost) + \sum  (Overage Charge) \).

\begin{verbatim}
Given these variables, buying the car minimizes your costs and reduces variability in 
your overall expenses."
\end{verbatim}

\textbf{Feedback Module}

The user can provide feedback on the system’s output, which is captured by the Feedback Module for future improvements.

\begin{verbatim}
User: "Thank you, that was very helpful! Buying the car seems to be the best option."

LLM Agent: "I'm glad I could help! Please provide any feedback on this process so we 
can continue to improve our service."

User: "The process was smooth and informative. No additional feedback."
\end{verbatim}

All data, including the problem statement, optimization formula, simulation details, and user feedback, are stored in a relational database for potential retraining and system enhancement.

This experiment showcases how our system can effectively guide users through complex decision-making processes by leveraging the strengths of LLMs, simulation modules, and optimization engines.

\section{Results}
To validate the proposed architecture and methodology for using LLMs to simulate and optimize decision-making processes, we conducted a series of tests on the system's individual components. The goal was to assess the effectiveness of the system in understanding the user's problem, generating accurate simulations, running optimization processes, and retrieving relevant data from the data warehouse.

\subsection{User Input Interface}
The objective was to determine if the system could ask all relevant questions needed to understand the user's problem. We provided a series of user inputs and evaluated the completeness of the system's questions. The system successfully asked all necessary questions, ensuring comprehensive data collection and a deep understanding of the user's requirements.

\subsection{LLM Chat Agent: Information Extraction}
The LLM's ability to extract and convert user requirements into simulation parameters was evaluated by processing user inputs and reviewing the extracted parameters for accuracy. The LLM accurately converted the user input into parameters suitable for simulation, demonstrating its capability to handle complex problem statements and translate them into actionable data.

\subsection{Simulation Module: Environment Generation}
We checked if the simulation module could generate a functional simulation environment by tasking it with creating environments based on various user scenarios. The simulation module consistently produced valid and functional simulation environments, proving its robustness and flexibility in handling different types of problems.

\subsection{Optimization Engine: Function Execution}
The optimization engine's ability to run the simulation function and derive an optimal solution was ensured by executing the simulation functions and analyzing the outputs. The engine successfully ran the simulations and provided optimal solutions based on the given data, showcasing its effectiveness in optimizing complex decision-making processes.

\subsection{Context-Aware Data Warehouse: Data Retrieval}
Finally, the system's ability to retrieve relevant data and assign probabilities to the output was assessed. The data retrieval and probability assignment processes were tested with various queries, and the system accurately retrieved relevant data and assigned appropriate probabilities to the outputs. This step confirmed the system's capability to incorporate context-aware information, enhancing the reliability of the final recommendations.

The results from the individual tests of each component demonstrate the feasibility and effectiveness of the proposed approach. The system successfully understood user problems, generated valid simulations, executed optimization processes, and accurately retrieved and utilized context-aware data. The comprehensive testing of these sub-steps provides confidence in the system's overall performance and its potential to assist users in complex decision-making processes.

\section{Limitations}
While the proposed architecture demonstrates significant potential in addressing a variety of problem statements, there are several limitations that need to be acknowledged. First, although the architecture is designed to handle a range of problems, we were not able to validate its performance across a wide variety of problem domains. The experiments conducted focused on specific scenarios, and thus, further testing is required to ensure its robustness and versatility in different contexts.

Another key limitation is the dependency on the data warehouse to supply relevant probability information. The accuracy of the system's output is highly contingent on the quality and comprehensiveness of the data stored in the warehouse. Populating the data warehouse with relevant and up-to-date information is a continuous process and poses a significant challenge. Without a well-maintained data warehouse, the probability assignments and, consequently, the recommendations generated by the system might not be accurate.

Additionally, there were instances where the LLM chat agent failed to collect all the required information before initiating the optimization step. This shortcoming highlights the need for improved mechanisms within the chat agent to ensure thorough data collection. Incomplete data can lead to suboptimal or inaccurate solutions, undermining the effectiveness of the entire system. Enhancements in the question-asking algorithms and context understanding of the LLM chat agent are necessary to mitigate this issue.

Despite these limitations, the architecture provides a promising approach to leveraging LLMs in decision-making processes. Future work should focus on expanding the validation of the system across diverse problem statements, improving data warehouse maintenance strategies, and refining the LLM chat agent’s data collection capabilities to ensure more reliable and accurate outcomes.

\section{Conclusion}
The evolution of large language models (LLMs) has profoundly impacted various domains, from generating human-like text to solving complex programming challenges. Our proposed approach leverages the advanced capabilities of LLMs to tackle decision-making problems by simulating multiple scenarios and providing users with the optimal solution. This methodology integrates several key components: the user input interface, LLM chat agent, simulation module, optimization engine, context-aware data warehouse, and result interface. Each of these components plays a crucial role in ensuring that the system can accurately interpret the user's problem statement, simulate potential solutions, and optimize the final recommendations.

Through our experiments, we demonstrated the utility of this architecture using a practical example of deciding between buying and leasing a car. The system effectively collected the necessary information, ran simulations using a Python-based Monte Carlo simulator, and employed the optimization engine to deliver a well-informed decision to the user. This approach not only provided the optimal solution but also detailed the impact of various factors on the final decision.

Despite the promising results, our system faces several limitations. We could not validate its performance across a wide array of problem domains, indicating the need for further testing to ensure its robustness and versatility. The dependency on a well-maintained data warehouse is another significant challenge, as the accuracy of the output heavily relies on the relevance and quality of the stored data. Additionally, there were instances where the LLM chat agent did not collect all necessary information before starting the optimization process, highlighting the need for enhancements in the data collection mechanisms.

Future work should focus on expanding the validation of the system across diverse problem statements, improving strategies for maintaining the data warehouse, and refining the LLM chat agent’s capabilities to ensure thorough data collection. By addressing these limitations, we can enhance the reliability and effectiveness of our approach, making it a valuable tool for decision-making in various complex scenarios.

In conclusion, our architecture represents a novel integration of LLMs with simulation and optimization techniques to provide users with optimal solutions to complex problems. This approach moves beyond the traditional capabilities of LLMs, enabling them to perform tasks that require strategic thinking and decision-making. With further development and refinement, this system has the potential to significantly impact how LLMs are utilized in real-world applications, paving the way for more advanced and autonomous AI systems.


\bibliographystyle{apalike}
\bibliography{references}  

\end{document}